\pdfoutput=1
\documentclass[10pt, a4paper]{article}
\usepackage{lrec-coling2024} % this is the new style
\usepackage{covington}
\usepackage{booktabs}
\usepackage{multirow}
\usepackage{xspace}
\usepackage{amsmath}
\usepackage{booktabs}
\usepackage{microtype}
\usepackage{nicefrac}

\hypersetup{breaklinks=true}
\usepackage{comment}
\makeatletter
\def\cl@chapter{\@elt {theorem}}
\makeatother
\usepackage[capitalise,noabbrev,nameinlink]{cleveref}

\title{It's Difficult to be Neutral -- Human and LLM-based \\Sentiment Annotation of Patient Comments}

\newcommand\Norec{{NoReC}\xspace}

\newcommand\NorecFine{{NoReC$_\text{\textit{fine}}$}\xspace}

\newcommand{\norex}[1]{\textit{#1}}
\newcommand{\eng}[1]{`#1'}

\newcommand\fen{{F$_{1}$}\xspace}

\crefname{equation}{Example}{Examples}
\name{Petter M{\ae}hlum$^1$, David Samuel$^1$, Rebecka Maria Norman$^2$, Elma Jelin$^2$,\\
{\bf \large{\O}yvind Andresen Bjertn{\ae}s$^2$, Lilja {\O}vrelid$^1$, Erik Velldal$^1$}} 

\address{$^1$University of Oslo, $^2$Norwegian Institute of Public Health, \\
         $^1$\{pettemae, davisamu, liljao, erikve\}@ifi.uio.no, \\ $^2$\{rebecka.norman, elma.jelin,oyvindandresen.bjertnaes\}@fhi.no,}

\abstract{
Sentiment analysis is an important tool for aggregating patient voices, in order to provide targeted improvements in healthcare services. A prerequisite for this is the availability of in-domain data annotated for sentiment.  
This article documents an effort to add sentiment annotations to free-text comments in patient surveys collected by the Norwegian Institute of Public Health (NIPH). 
However, annotation can be a time-consuming and resource-intensive process, particularly when it requires domain expertise. We therefore also evaluate a possible alternative to human annotation, using large language models (LLMs) as annotators. We perform an extensive evaluation of the approach for two openly available pretrained LLMs for Norwegian, experimenting with different configurations of prompts and in-context learning, comparing their performance to human annotators. We find that even for zero-shot runs, models perform well above the baseline for binary sentiment, but still cannot compete with human annotators on the full dataset. 
 \\ \newline  }

\begin{document}

\maketitleabstract

% These are maybe the most important points, although not limited to this?
    
    %Linguistic resources, data and tools for language technologies focusing on consumer health
    
    %Processing and annotation platforms
    %Synthetic data generation and data augmentation.

\section{Introduction}

The Norwegian government has a long tradition of collecting data on patient experiences in the form of surveys, and recently this has also included of unstructured free-text comments. The application of sentiment analysis (SA) to these texts is expected to provide valuable information on patient experiences, which can then be used to improve care at both district and national levels. 

This paper documents a large-scale annotation effort to add comment- and sentence-level polarity to patient feedback, representing a collaboration between NLP researchers and health professionals. Specifically, we target patient comments on experiences with General Practitioners and Special Mental Healthcare.
In addition to presenting the annotation guidelines and an analysis of the resulting dataset, we also include experimental results on augmenting the human annotations with predictions by pretrained large language models (LLMs). Using two recently released generative LLMs for Norwegian based on the T5 and Mistral architectures \citep{10.5555/3455716.3455856, jiang2023mistral}, we present results for different prompts combined with zero- and few-shot learning. We also compare and discuss the differences in error types made by human annotators and the models. 
 
Due to privacy concerns and the sensitive nature of patient feedback, the underlying text material can unfortunately not be openly distributed, but we publish the prompts, the class distributions and the annotation guidelines.\footnote{\url{https://github.com/ltgoslo/Sentiment-Annotation-of-Patient-Comments/}}
%While privacy concerns disallow us from releasing the data, all prompts and experimental results will be made public.
%More concretely, we try to answer (1) Which performance can be seen by Norwegian LLMs on patient feedback data, and (2) To what extent and in what way are the model predictions different from human annotations.

\section{Background and Motivation}

\paragraph{Importance of patient feedback} Systematic reviews of the literature show that positive patient experiences are associated with better patient safety, better effectiveness, higher levels of adherence, and lower healthcare utilization \citep{Anhang_Price2014-ln, Doylee001570}. %The Norwegian government has ordered the patients’ health service, including stronger involvement of patients in health care decisions and in the development and evaluation of services.\footnote{Nasjonal helse- og sykehusplan 2020-23 (2019) Helse- og omsorgsdepartementet.}. 

An important patient-oriented tool at the national level is the national system for measuring patient experiences. The purpose of the system is to systematically measure patient experiences with health services, as a basis for quality improvement, management of health services, free patient choice, and public accountability. To underpin these goals, quantitative results from surveys are produced and presented at different health care levels, e.g., results for hospitals, for health regions, and results at the national level.

\paragraph{NIPH Surveys} The Norwegian Institute of Public Health (NIPH) has conducted many national patient experience surveys. All surveys include one or more open-ended questions in which patients are encouraged to write about their experiences with the health service, which is equivalent to tens of thousands of comments each year. These comments present a rich data source on health service evaluation \citep{Grob2019-do, rivas2019automated}, but are mostly unused due to the time and resources needed to analyze them. NIPH's current approach is to conduct manual content analysis of a random sample of 500 comments in each survey and report main findings at the national level alongside quantitative results. Furthermore, providers might get access to the data for their patients, but most providers lack the competence, systems and resources to analyze qualitative data. This means thousands of free text comments from each survey are excluded from further analysis and consequently also from provider-level reports. This exclusion is problematic from an ethical point of view, but also because these types of data at lower levels are highly valued by providers \citep{Riiskjaer2012-uh} and are well-suited for use in quality improvement \citep{Grob2019-do}. Free-text comments from surveys have the potential to nuance the quantitative data. For instance, a substantial proportion of patients with the highest quantitative scores describe negative experiences in free-text comments \citep{iversen}, indicating that these could be used for differentiating patients at the higher end of the scale. Thus, one can expect added value of quantitative indicators at the provider-level based on the qualitative feedback of text comments.

%Our experience with free-text comments from surveys is that they nuance the quantitative data and add information.

Therefore, there is a clear need for an innovative and highly efficient method for analyzing large amounts of patient comments. In this paper we describe the first steps towards the automatic analyses of Norwegian free-text comments from patients. Feedback reports with results of free text analysis at the provider-level will make them more relevant and actionable for clinicians and managers who want to improve quality \citep{Grob2019-do, Riiskjaer2012-uh}, thus possibly also strengthening the patient's voice in quality improvement. Alleviating the workload and costs associated with annotating data for these systems constitutes an important step in this direction.

%Originaltekst fra FHI
%This project will develop and evaluate machine learning tools for automatic analyses of free-text comments from patients. The development and subsequent application of state-of-the-art Natural Language Processing to patient comments will strengthen the patient voice in healthcare evaluations by analysing all patient comments (not just a random sample), which in turn facilitate the inclusion of such data in provider-level reports. Feedback reports with results of free text analysis at the provider level will make them more relevant and actionable for clinicians and managers aiming to improve quality \citep{Grob2019-do, Riiskjaer2012-uh}, thus possibly also strengthening the patient voice in quality improvement. 

\section{Previous work}

%Petter,
%Ta utgangspunkt i Elmas powerpoint
\paragraph{NLP for Patient Feedback} \citet{Khanbhai2021} present a systematic review on the application of NLP and machine learning techniques to patient experience feedback. It shows that 80\% of the surveyed studies applied language analysis techniques to patient feedback from social media sites followed by structured surveys. These studies include work based on both supervised and unsupervised learning for text and sentiment analysis (SA). To provide data for supervised SA, previous work relies heavily on manual classification of a subset of data by themes and sentiment \citep{alemi2012, bahja2016, Doing-Harris2017-hq, greaves2013, Hawkins404, huppertz, Wagland604,zafra}. In previous work, comments extracted from social media have also been analyzed using an unsupervised approach; however, free-text comments from surveys are typically analyzed using supervised machine learning \citep{Khanbhai2021}. \citet{Khanbhai2021} discusses that comprehensive reading of all comments within the dataset remains the `gold standard’ method for analyzing free-text comments, and that this is currently the only way to ensure that all relevant comments are coded and analyzed, demonstrating that language analysis using an ML approach is only as good as the dataset used to inform it. Other studies recently published in this area are all examples of how NLP and SA can be used to make the information more accessible and usable in various quality improvement initiatives, for example, using dashboards, pipelines, and visualization \citep{alexander2022, khanbai2022, Rohde2022, vanBuchem2022}.

\paragraph{Norwegian Sentiment Analysis} To our best knowledge, there has been no previous work on sentiment analysis (SA) for free-text patient feedback, or for any user-generated text in Norwegian. The bulk of previous work on Norwegian SA in general relies on the \Norec dataset of multi-domain reviews collected from various news sources \citep{velldal-etal-2018-norec}. Based on an annotated subset, \citet{ovrelid-etal-2020-fine} have published a fine-grained SA corpus (\NorecFine) along with annotation guidelines, on which we partly base our manual annotation effort, further described in \cref{section:annotation}. \NorecFine contains annotations for fine-grained sentiment: annotating the opinion holder, target and polarity. In addition to this, previous work has focused on entity-level aggregation of SA annotation \citep{ronningstad-etal-2022-entity}, and also on improving existing models using data augmentation based on a masked language model \citep{kolesnichenko-etal-2023-word}. %The task of (sentence-level and targeted) SA is further part of NorBench \citep{samuel-etal-2023-norbench}, a benchmark suite for various Norwegian language models, based on benchmarks derived from \NorecFine. NorBench focus mainly on encoder models like BERT, assessing their capabilities after fine-tuning on task-specific datasets.
It's worth noting that, being written by professionals reviewers, the documents in NoReC do not contain many of the features that are typical of user-generated data, as in our patient comments, apart from the obvious differences in domain.

\paragraph{Zero/Few-Shot Evaluation} In this work, the main focus will be on decoder-based, or generative, language models and the evaluation of their capabilities as sentiment annotators. These types of models have been shown to perform well in zero-shot or few-shot settings \citep{brown2020language, wei2022finetuned} with limited annotated data. 
Given the possibility of accessing these models via natural language prompts, they are arguably also easier to use than traditional models, especially in consideration of health professionals who may not have programming experience. 
We further focus exclusively on openly available models that can be run locally and do not risk leaking of sensitive data via an API to a proprietary service. This is crucial in our data setting, where surveys are considered sensitive data.

\section{Annotation}
\label{section:annotation}

%The data were anonymized for project internal use

\paragraph{Scope and Sources} Our data comes from free-text comments from the NIPH patient experience surveys described above.  The surveys cover various different domains, but we focus on two subsets of data: experiences with General Practitioners (GPs) and evaluation of Special Mental Healthcare (SMH). The data from these surveys has been the focus of earlier research \citep{Kjollesdal2020-wz,Iversen2022-ju}, but the use of machine learning methods on the free-text comments is new. While the domain (health-related) and the genre (user-generated text) in these surveys are quite different from the professional reviews found in the existing \Norec corpus, we adopt a similar annotation setup as that used in the \NorecFine \citep{ovrelid-etal-2020-fine} annotation effort. The original annotations were done at both comment- and sentence-level, with a three-way intensity scale, together with positive and negative sentiment, as well as explicit mention of neutral sentences, indicating that there is no expression of sentiment. Rather than using the full space of distinct labels allowed by this annotation scheme, in the experiments reported later we will only use four, corresponding to positive, negative, mixed, and neutral, ignoring the intensities.

\paragraph{Annotators} Annotation was performed by seven researchers in the health service. A set of annotation guidelines was devised based on \NorecFine, with some adaptations to the active domain. Certain aspects of \NorecFine, such as the distinction between various types of sentiment, and the precise delimitation of holders, targets and polar expressions were not carried over. The seven researchers annotated in rounds of 50 comments each. In addition to this, annotators received small sets of 20 comments for quality control and Inter-Annotator Agreement (IAA) calculations. These datasets were annotated without discussions, but the results were discussed in order to resolve any ambiguity or potential issues. %Due to work load constraints, the number of annotators was reduced towards the end of the annotation process. %Annotation is unequally distributed among the annotators, with the most annotated by one annotator being 900 comments, and the least 50 comments. 
%All annotators were familiar with the surveys that were evaluated, but annotators more familiar with either GPs or SMH were assigned mainly to these. % but as some were more familiar with certain parts, each annotator mostly worked on either GP data or SMH data. 

\subsection{Annotation Procedure}
The sentiment annotation was performed at both the comment-level and the sentence-level. 
The annotators marked polarity (positive, negative) and intensity (slight, standard, strong). The original texts were sentence-segmented and tokenized using Stanza \citep{qi2020stanza}. Manual inspection of the resulting data shows that the tool provides accurate segmentation; however, cases such as emoticons (`\texttt{:-D}', etc.) are sometimes split erroneously. 
Sentences containing both positive and negative sentiment were annotated separately for both polarities. Sentences containing no polarity were marked as neutral. 

%in the \texttt{.xlsx} format. Polarity was annotated using a set of letters: \texttt{s}, \texttt{ss}, and \texttt{sss}, indicating \textit{slight}, \textit{standard}, and \textit{strong} intensity, in either a column for positive or a column for negative sentiment. A sentence could thus contain both positive and negative sentiment, to different degrees. A sentence containing no sentiment was marked as \texttt{b} (\textit{blank}). 

%Ikke så farlig akkurat hvordan det ble gjort og hvilke labels de brukte, eller hvilke verktøy det er.

Comments were generally annotated using the same set of guidelines as for sentences, but comment-level polarity was scored based on a general impression of the comment as a whole, not just an aggregate of the sentiment of the sentences it contains. Annotators placed special weight on how the actual service (GP, SMH) was evaluated when assigning labels at the comment-level. Basic comment statistics are reported in \cref{tab:corpus}.

\begin{table}[]
    \centering
    \begin{tabular}{@{}lrrr@{}}
        \toprule 
        \textbf{Type} & \textbf{Total} & \textbf{SMH} & \textbf{GP} \\
        \midrule 
        Comments & 2\,250 & 1\,050 & 1\,200 \\
        Sentences per comment &3.4 & 3.5 & 3.3 \\
        Words per sentence &14.8 & 16.0 & 13.8 \\
         \bottomrule  
    \end{tabular}
    \caption{Number of comments, and average number of sentences per comment and words per sentence. We see that feedback in psychiatric care tends to be longer in both measures.}
    \label{tab:corpus}
\end{table}

%\paragraph{Irrelevant information} was indicated by \texttt{i}. This signalled that the content of a sentence contained some sentiment that was not relevant to the main target. All sentences were annotated as comments on GP or PHV, and a sentence marked as \texttt{i} would contain something not realated to this. In practice, these sentences should not be automatically aggregated to give an indication of patients' sentiment towards a practice, but in this study we have not singled them out.

\paragraph{Examples} The comments in the dataset vary in terms of how sentiment is expressed, and sentiment expressions can contain direct evaluations, as we see in \cref{ex:strong}, where we find the strong positive adjective \norex{fantastisk} \eng{fantastic} describing hospital employees, as well as the adjective \norex{fin} \eng{nice} describing the patient's stay. 

{\small
\begin{covexample}
\gll Fantastiske ansatte som har gjort oppholdet så fint .
fantastic employees who have made stay.the so nice .
\glt‘Wonderful employees that have made my stay so nice.’
\glend
\label{ex:strong}
\end{covexample}
}

\noindent
In \cref{ex:slight} on the other hand, we see a more indirect evaluation of the treatment. The suggestion that a video consultation is useful is interpreted as a slightly negative evaluation of the health service in question.

{\small
\begin{covexample}
\gll Kanskje kunne videokonsultasjon være til nytte ved en slik livssituasjon ?
Maybe could video.consultation be to use by a such life.situation ?
\glt‘Perhaps a video consultation could be of use in this kind of life situation?’
\glend
\label{ex:slight}
\end{covexample}
}

\noindent
The data also contain several neutral examples --   typically patients reporting on their own health situation, as in \cref{ex:blank}, or general descriptions. %These also include sentences that do not have discernible content, such as sentence fragments.

{\small
\begin{covexample}
\gll I tillegg bruker jeg øyedråper mot høyt trykk ( grønn~stær ) .
In addition use I eye.drops against high pressure ( glaucoma ) .
\glt‘In addition, I use eye drops for high pressure (glaucoma).’
\glend
\label{ex:blank}
\end{covexample}
}

%\paragraph{Comparative} sentences were marked as \texttt{K} (for Nor. \textit{komparativ} `comparative'). This was done as annotators reported that comparative sentences were challenging, and this is also know from earlier literature. The tag was included to provide the possibility of seeing if these sentences are more difficult. 

%\paragraph{Questionnaire-related} sentences were marked as \texttt{q}. These were used to separate out the sentiment towards the questionnaire in which the patients write their feedback. All questions related to the reflections or comments of the patients on this form were marked with \texttt{q}. 

%Annotators would add annotations in columns, and could also mark sentences as irrelevant. Early discussions also lead to the inclusion of a marker for comparative expressions, as these were found to be difficult to interpret, and earlier sentiment research has also mentioned them as problematic (Liu). 

\paragraph{Further Discussions} The annotators had weekly meetings both with each other, and with researchers working on sentiment analysis, to discuss problematic cases and to update the guidelines so that they better reflect the choices taken. Some of the issues discussed include to what extent descriptive sentences can indicate the patients' opinion, and to what extent context should be taken into account. In cases where sentiment would be ambiguous without context, annotators were asked to use the full context of the comment, and also their knowledge of the field. In some cases, apparently negative words could indicate positive sentiment, as in cases where patients note that they get diagnosed with a serious illness, perhaps indicating that it was good that it was actually diagnosed, or that it reflects well on the GP who diagnosed them correctly, and not focusing on the negative aspect of having a disease in itself.

\subsection{Style and Variation} 
The data is notably different from the existing resources for Norwegian sentiment analysis. Being user-generated language, without editing, it contains some variation, both grammatical and stylistic. Certain terms pertaining to the domains, such as \norex{fastlege} \eng{general practitioner}, and \norex{opphold} \eng{stay} are naturally very frequent. We also observe a tendency for certain polar expressions to be very frequent, with examples including \norex{fornøyd} \eng{pleased} being among the top 20 most frequent words in the training set. A similar tendency is not observed in the \NorecFine data, where there is more variation and a weaker tendency for certain polar expressions to dominate. However, the texts notably do not contain much medical jargon, although some terms related to diseases and treatments. We believe the texts to be different from typical \textit{medical domain} writing. The texts further reflect the society to some degree: both official Norwegian written norms are represented, Bokmål and Nynorsk. There is also some English in the dataset. Finally, some comments show indications of lexical and syntactic patterns associated with learner language, and with spoken language in general. For example, subjectless sentences are relatively common, as in \cref{ex:subjectless}%(SMH, standard negative)
, where the subject of the verbs \norex{har} and \norex{kan} has been elided.  

{\small
\begin{covexample}
\gll Har vert innlagt før og kan trygt si at her burde noen gripe tak .
Have been admitted before and can safely say that here should some grab hold . 
\glt‘[I] have been admitted before and can safely say that someone should address the problems here.’
\glend
\label{ex:subjectless}
\end{covexample}
}

%Add some more dataset diagnostics

% The corpus cannot be made accessible due to its sensitive nature, but guidelines will be published, along with full model results. % kunne vi slippe et lite subset av veldig anonymiserte eksempler? 

%46 separate rounds of annotations, each containing 50 comments each, were added. All data are annotated by health professionals. 

% Number of annotators on different documents involved, so that we see if there is any difference.

%We now compare the results with the human annotators
%Annotators would annotate sets of 20 documents, 10 from GP and 10 from PHV, with regular intervals, and with regular meetings. Inter-annotator agreement was measured using Cohen's kappa, for all possible annotator pairs. 
%Initially, the human annotators had mediocre scores, with a marked difference between certain pairs of annotators, albeit on small datasets. Overall, the tendency is that more annotators disagree on negative sentences. The annotations were done both at comment and sentence level, but we will focus on sentence level in this paper. 

\subsection{Data Preparation and Splits}
Although the full dataset is annotated with information about intensity, we decided not to include these attributes for the experiments in the current study due to the added complexity. Sentences containing slight, standard or strong polarity are thus labeled as only belonging to one of the two polarity classes, \textsc{pos} or \textsc{neg}. Sentences containing both of these, in any polarity constellation, are labeled as \textsc{mix}, while sentences with no polarity are labeled \textsc{neut}. The distribution of these four classes in the dataset is reported in \cref{tab:classes}. As we can see, the class distribution is well balanced across the training and test data, which were split 50--50, motivated by the need to ensure the test set is large enough and the fact that we do not perform any training. The split was done making sure that both datasets contained a similar number of labels for each class. The sentences were randomly sampled for each class.   
We mainly focus our experiments on this four-class setup, because we consider this to correspond most closely to a realistic use-case and because distinguishing sentiment-containing data from neutral sentences is an important part of naturally occurring data. However, we also report results for a reduced 2-class version, discarding the \textsc{mix} and \textsc{neut} examples.

 %It was important for us to include this distinction, as the data these models will be used on will will contain classes, so removing a mixed class or neutral sentences would skew our results. Due to the limited size, and because we do not perform fine-tuning, the dataset was split 50-50 for train and test splits. The train split is used to extract few-shot examples.

\begin{table}[]
    \centering
    \begin{tabular}{@{}lrrrr@{}}
    \toprule
      & \textbf{\textsc{pos}} & \textbf{\textsc{neg}} & \textbf{\textsc{neut}} & \textbf{\textsc{mix}} \\
    \midrule
    Train & 1\,396 & 1\,753 & 476 & 220 \\
    Test & 1\,396 & 1\,755 & 477 & 220 \\%[0.5em]
    Total & 2\,792 & 3\,508 & 953 & 440 \\
        \bottomrule
    \end{tabular}
    \caption{Class distribution in the dataset for Positive, Negative, Neutral and Mixed sentences. The high similarity in counts is due to all classes being weighted when creating the splits, making sure both splits were as similar as possible.}
    \label{tab:classes}
\end{table}

%The commments were split and tokenized using Stanza (cite). 

\subsection{Human Inter-Annotator Agreement}

\paragraph{First Round} Initially, two test rounds of annotation were performed, in order to evaluate the guidelines, the experience of the annotator, and their agreement. Following an initial set of test annotations, smaller sets of 20 comments were annotated with regular intervals to judge the annotators' progress. Although the results from the first of these sets showed a large variance between annotators, with kappa agreement scores for annotations including intensity varying from very low at 0.21 to as high as 0.76. We note that much of the disagreements stem from the annotations of intensity. However, when considering to what extent annotators agree on the two polarity classes and neutral sentences, ignoring intensity, the lowest kappa score was 0.62, and the highest 0.93.   %providing the 3-class scores shown in 

\paragraph{Second Round} Following the results from the first of the 20-comment annotations, the annotators discussed various specific cases of disagreement. Topics of discussion included to what extent certain expressions exhibit sentiment, and annotators were asked to give justification for their annotations, which were then discussed before consensus was made, if possible. After a series of discussions, a new IAA dataset was annotated to show to what extent these discussions affected the agreement among annotators, giving the results in
\cref{tab:compared}, not including intensity, where we see that scores have improved markedly, showing that humans agree on which sentences are positive, negative and neutral. There is notably more disagreement for the negative polarity than for the positive. We also observe that there are often more than one annotator disagreeing for negative polarity, indicating greater variation and uncertainty for negative labels. One annotator annotated for the project, but did not partake in these two IAA-rounds.%This applies for both these rounds, even though there is some improvement in the second round.

\begin{table}[]
    \centering
    \begin{tabular}{@{}lllllll@{}}
    \toprule
 &  \textbf{A1}    &  \textbf{A2}   &  \textbf{A3}   &  \textbf{A4}    &  \textbf{A5}   &  \textbf{A6}     \\
 \midrule
A1 &  1.0    & 0.96   & 0.95   & 0.95   & 0.96   & 0.95   \\
A2  &  0.96   & 1.0    & 0.95   & 0.92   & 0.95   & 0.92   \\
A3  &  0.95   & 0.95   & 1.0    & 0.95   & 0.95   & 0.93   \\
A4  &  0.95   & 0.92   & 0.95   & 1.0    & 0.92   & 0.93   \\
A5  &  0.96   & 0.95   & 0.95   & 0.92   & 1.0    & 0.92   \\
A6  &  0.95   & 0.92   & 0.93   & 0.93   & 0.92   & 1.0    \\
\bottomrule
    \end{tabular}
    \caption{Inter annotator agreement scores for positive, negative and neutral sentiment from the second round of IAA-annotations. One annotator who annotated for the project did not partake in these specific rounds.}
    \label{tab:compared}
\end{table}

\section{LLM-Based Annotation}
For free-text comment analysis, finding human annotators can be difficult. Not only are there few to choose from who have the necessary background knowledge to interpret the context correctly, annotation work can be expensive and time-consuming, placing additional economic strain on health services researchers or other personnel annotating data. We want to explore whether newer models could aid in this effort. Therefore, we compare the zero-shot and few-shot performance of two Norwegian LLMs with the annotations of six healthcare professionals.

\subsection{Language Models}

In particular, we evaluate predictions of one relatively small but instruction-finetuned model, \textsc{ChatNorT5}, and of a larger model that has not been finetuned on any downstream tasks, \textsc{NorMistral}.

\paragraph{\textsc{ChatNorT5}} This model is an instruction-finetuned version of \texttt{nort5-large}, an 808-million-parameter Norwegian encoder-decoder language model \citep{samuel-etal-2023-norbench}.%
%\footnote{\ttfamily\href{https://huggingface.co/ltg/nort5-large}{https://huggingface.co/ltg/nort5-}\\ \href{https://huggingface.co/ltg/nort5-large}{large}}
\footnote{\url{https://huggingface.co/ltg/nort5-large}}
By itself, NorT5 is pretrained on masked language modeling \citep{10.5555/3455716.3455856}, therefore, we further finetune it on instructions (via causal language modeling), to turn it into a generative model capable of predicting sentiment labels in zero-shot or few-shot settings. To make the evaluation more realistic, we train on a general set of instructions, not specifically focusing on sentiment analysis. We use a collection of Alpaca-like datasets \citep{taori2023alpaca} and translate them from English to Norwegian Bokmål with OPUS-MT \citep{TiedemannThottingal:EAMT2020}.\footnote{UltraChat \citep{ding2023enhancing}: \url{https://huggingface.co/datasets/stingning/ultrachat}, ShareGPT: \url{https://huggingface.co/datasets/philschmid/sharegpt-raw}, WizardLM: \url{https://huggingface.co/datasets/WizardLM/WizardLM_evol_instruct_V2_196k} and SODA \citep{kim2022soda}: \url{https://huggingface.co/datasets/allenai/soda}.} In total, we have translated 287k conversations and then finetuned the model for one epoch. One conversation consists of multiple query-response turns, and the model is trained to produce a gold response (using its decoder part) given all previous turns (provided to the encoder part). 

\paragraph{\textsc{NorMistral}} In addition to the instruction-tuned model, we also test an openly available Norwegian language model called \texttt{normistral-7b-warm}.\footnote{\url{https://huggingface.co/norallm/normistral-7b-warm}} This model has been continually pretrained from the Mistral 7B model \citep{jiang2023mistral} and has been shown to perform well in zero-shot and few-shot evaluations, even without instruction-tuning.

\subsection{Likelihood Scores}

As using the generated output from a causal LLM can lead to difficulties in mapping to the gold classes, we only consider a fixed set of possible responses for each prompt. We follow \newcite{brown2020language} and formulate sentiment analysis as a task of choosing one prompt completion out of a limited number of other possible completions, based on their \textit{likelihood scores}. Both evaluated language models can output $P(s_i\,|\,s_{0\,:\,i})$, the estimated probability of producing a subword $s_i$ given the previous subwords $s_{0\,:\,i}=s_0\dots s_{i-1}$. We use this ability to test three ways of calculating a likelihood score of a completion $c=(c_0\dots c_n)$ given a query $q$:
\begin{enumerate}
    \item $\mathcal{L}_1(c\,|\,q) = \sum_{c_i \in c}{logP(c_i\,|\,q, c_{0\,:\,i})}$,
    \item $\mathcal{L}_2(c\,|\,q) = \nicefrac{1}{n} \cdot \sum_{c_i \in c}{logP(c_i\,|\,q, c_{0\,:\,i})}$,
    \item $\mathcal{L}_3(c\,|\,q) = \nicefrac{1}{n_{\text{char}}} \cdot \sum_{c_i \in c}{logP(c_i\,|\,q, c_{0\,:\,i})}$.
\end{enumerate}

\noindent
The first formula calculates the actual estimated log-probability of $c$ given $q$; however, in practice, this formulation tends to overestimate the likelihood of short sequences. Therefore, we try to normalize the likelihood by the length of completion $c$ -- the second formula normalizes by its number of subwords, $n$, and the third formula by its number of characters, $n_{\text{char}}$.
% Three possible aggregations were selected for ChatNorT5 ...%This section might be rewritten as a part of the section on ChatNOrT5. 
% (Should mention that ChatNorT5 has seen some sentiment data)

\subsection{Prompting}
%Our experiments involve calculating the likelihood of a specific reply indicating some class, given a specific prompt. 
As there is little earlier work on prompts for sentiment analysis for Norwegian, we based our initial prompts on the existing sentiment-related prompts in the FLAN dataset \cite{wei2022finetuned}. The FLAN dataset contains four datasets: IMDB \citep{maas-etal-2011-learning}, Sent140 \citep{sentiment140}, SST-2 \citet{socher-etal-2013-recursive} and Yelp.\footnote{\url{https://course20.fast.ai/datasets.html}} FLAN includes several sets of English SA prompts for each of these, and they were manually translated into Norwegian.%, with some modification. 

\paragraph{Prompt Variation} The prompts from FLAN contain variation in terms of multiple choice variation, differences in formality, as well as different near synonyms, and the words used to refer to the text itself (\textit{the preceding}, \textit{this}). We aimed at keeping some variation, but discarded multiple choice questions and informal variants, and did not experiment with synonyms.

%Some parts of the prompts are repeated for several of the datasets, such as \textit{``What is the sentiment of this..."}, but at times with small variations. Types of questions include direct questions (\textit{``What is the sentiment..."}), also with various types of multiple-choice variations (\textit{``choose"}, \textit{``select"}). The prompts included various references to the text to be predicted (\textit{``the following"}, `\textit{`this"}, \textit{``the preceding"}), variation in the relationship between sentiment and the text (\textit{``of"}, \textit{``embodied by"}). In addition, the dataset contains several prompts to generate text. We have not used these generative prompts in our experiments. Finally, some sentences include less formal and more subjective phrasings (\textit{``well"}, \textit{``How would"}, \textit{``what is... like?"}).

\paragraph{Prompt Filtering} Among all Norwegian translations of prompts, only natural-sounding sentences were considered. As we calculate likelihoods based on a certain reply given a prompt, we also wanted to keep the number of possible replies low. Some sentences were discarded due to requiring very different replies. We also wanted to be able to compare the prompts with each other, and therefore excluded sentences that would force us to expand our number of prompts drastically. 
%and therefore excluded sentences that would require  Other reasons for discarding include difficulty in providing natural sounding corresponding sentences, and some sentences were discarded in order to not have too many confounding variables in the remaining prompts, as we wanted to be able to control for the results of certain types of variation. Prompts were also selected based on how they could be answered, as we wanted the answer space to be a bit smaller.
FLAN-prompts not related to sentiment classification were excluded. The resulting base prompts are shown in \cref{tab:sentences}.

\begin{table*}[] %remove *
    \centering
    \small
    \begin{tabular}{@{}cll@{}}
    \toprule 
    \textbf{ID} & \textbf{Norwegian} & \textbf{English} \\
    \midrule 
1 & Hvordan er sentimentet til denne setn.? & How is the sentiment of this sent.? \\
2 & Hva er sentimentet til denne setn.? & What is the sentiment of this sent.? \\
3 & Hvordan vil du beskrive sentimentet til denne setn.? & How would you describe the sentiment of this sent.?\\
4 & Beskriv sentimentet i denne setn..  &  Describe the sentiment in this sent..\\
5 & Ville du sagt at denne setn. er positiv eller negativ?  & Would you say that this sent. is positive or negative?\\
6 & Vil du si at denne setn. er positiv eller negativ? & Would you say that this sent. is positive or negative?  \\
7 & Er sentimentet i denne setn. positivt eller negativt? & Is the sentiment in this sent. positive or negative?\\
8 & Oppfatter du denne setn. som positiv eller negativ?  & Do you see this sent. as positive or negative? \\
9 & Er denne setn. positiv eller negativ?  & Is this sent. positive or negative?\\
\bottomrule 
    \end{tabular}
    \caption{The 9 base prompts, and their English translations. Note that the translations here are back-translations. sent.=sentence, setn. =setningen.}
    \label{tab:sentences}
\end{table*}

\paragraph{Prompt Expansion} These base prompts were then expanded to create 48 prompts. Each resulting prompt is given a 4-number code based on the kind of modification it received. The first number (1-9) indicates the base prompt from \cref{tab:sentences}. The second number indicates whether the test set sentence comes after (1) or before (2) the prompt. The third number indicates whether the question has no mentions of any of the four classes (0), mention of positive and negative (2), or all four (4). The final number indicates whether the word \norex{positiv} comes before (2) or after (1) the word  \norex{negative}. %were switched (2) or not (1). % and categorized. The original 9 were split into basic, with no leading questions, and binary, asking for either \textit{positive} or \textit{negative}. The test data sentences can come before or after (reverse). Then, the 5 binary prompts were extended to explicitly mention the four classes (\textit{positive}, \textit{negative}, \textit{mixed} and \textit{neutral}), also with the sentence either before or after. Finally, we introduced a simple permutation of the words "positive" and "negative" in both the binary and multiclass prompts. 
%We did not experiment with further permutations due to the exploding number of prompts. 
We give an example of prompt \texttt{8-2-4-2} in \cref{ex:promptet}. The id means that it is based on prompt 8, has the target sentence before the prompt, has 4 classes, and has \textit{negativ} before \textit{positiv}.

% OBS! 1 means behind, 2 means before
%\begin{itemize}
%    \item (1--9): Base prompt ID
%    \item (1--2) sentence behind or before
%    \item (0,2,4): zero, binary or multiclass
%    \item (1--2) unpermuted vs. permuted
%\end{itemize}

{\small
\begin{covexample}
\gll     Oppfatter du denne setningen som positiv, negativ, blandet eller nøytral?"
Consider you this sentence.the as positive, negative, mixed or neutral?
\glt‘Do you consider this sentence positive, negative, mixed or neutral?’
\glend
\label{ex:promptet}
\end{covexample}
}

\paragraph{Possible Replies to Prompts} 
%When assessing the capabilities of generative models, it would have been possible to evaluate based on the generated text and try to device post-processing methods that map the generated output to a class, and use that as annotation output. However, this can be difficult, and initial experiments show that the output from ChatNort5 is somewhat unpredictable. It has been shown that using likelihood scores can increase the models' performance, and since this removes the difficult aspects of dealing with generated output, we opt for using a set of possible answers, for which likelihoods are calculated. 
%Each prompt is combined with a limited set of possible answers, but we still examine some variation of these answers; this can be seen in \cref{ex:pos1,ex:pos2}.
% First, a full sentence reply to the sentence was created (\ref{ex:pos1} as a reply to a prompt such as \textit{Vil du si at denne setningen er positiv eller negativ?} `Would you say that this sentence is positive or negative.'. Then, the subject noun was reduced to a pronoun as in(\ref{ex:pos2}), or reduced to only the class label itself.
%finally only the keyword was left (\ref{ex:pos3}). Some alternative versions were added for the neutral and mixed classes. 
Each prompt is combined with a limited set of possible answers. Much of the variation in these replies comes from the two main classes of answers, one with the word \textit{setningen} `the sentence' and one containing \textit{sentimentet} `the sentiment', which require masculine (\textit{setningen})  and neuter (\textit{sentimentet}) agreement, respectively. We introduce versions of replies that have a pronoun (\textit{den}, `it') instead, and finally versions where only the class is mentioned in the reply.
In total, there are 16 alternatives per prompt, of which two have no difference depending on grammatical gender, giving 30 different replies. Each possible reply was associated with its suitable prompt data. A prompt file containing all possible prompts with all possible replies for each prompt was the basis of our experiments.

%We believe that the reason why the model does well on this is partially because it has been trained on this type of data. These sentences are not necessarily natural, in the sense that general native speakers ask questions like this.

Examples\ref{ex:pos1}, \ref{ex:pos2} and \ref{ex:pos3} show how the answers to a prompt vary in syntactic structure and content. Example \ref{ex:pos1} shows a full sentence referring back to the sentence (or sentiment), while in \ref{ex:pos2} it is substituted by a pronoun, and finally in \ref{ex:pos3} both the pronoun and the verb is elided to provide a minimal answer. 

\begin{covexample}
\gll Setningen er positiv .
the.sentence is positive .
\glt‘The sentence is positive.’
\glend
\label{ex:pos1}
\end{covexample}

\begin{covexample}
\gll Den er positiv .
it is positive .
\glt‘It is positive.’
\glend
\label{ex:pos2}
\end{covexample}

\begin{covexample}
\gll Positiv .
positive .
\glt‘Positive.’
\glend
\label{ex:pos3}
\end{covexample}

\section{LLM Experiments and Results}
We performed experiments for both zero-shot and few-shot set-ups for both \textsc{ChatNorT5} and  \textsc{NorMistral}. We look at both a 4-class representation and a reduced binary representation, which corresponds roughly to cases where we would expect low and high agreement, respectively.
To evaluate the binary dataset, we simply do not evaluate the model output on the neutral and mixed labels, and we limit the evaluation by only investigating the predicted likelihoods for the replies mapping to the classes \textsc{pos} and \textsc{neg}. The results are compared to a simple bag-of-words Naive Bayes model baseline.

\paragraph{Naive Bayes Baseline}
We set up a simple Naive Bayes model using the Natural Language ToolKit (NLTK) Python library \citep{bird2009natural}, using the entire vocabulary of the train set, removing the 20 most common words. With this baseline, we achieve a macro \fen score of 41.0 for the four-class problem, higher than the random baseline of \fen 22.0 For the binary setup we achieve a quite high \fen score of 79.0  compared to the baseline of 50.0 

\begin{comment}

#Binary
Accuracy 0.6792452830188679
Macro F1 0.5848479655756745
Macro F1 average 5 runs 0.47
#four class
Accuracy 0.559748427672956
Macro F1 0.3914260326819231
Macro F1 average 5 runs 0.22
FHI four class
Accuracy 0.6522869022869023
Macro F1 0.4095548149414446
Macro F1 average 5 runs 0.22
\end{comment}

\paragraph{Experimental Setup} Given the document with all 48 prompts and 16 alternatives per prompt, we estimated the likelihoods for each sentence in the test set. %These initial zero-shot evaluations were done both to select good prompts,which could then be used for few-shot learning. 
For a given sentence in the test set, there are 16$\times$3  likelihood measures. For each of the three different likelihood scores, we selected the maximum across these 16, mapped the reply alternatives to one of the four classes, and treated that as the predicted value of that sentence. %One important difference between the models is that while \textsc{ChatNorT5} is instruction-finetuned, \textsc{NorMistral} is not. We still have the possibility to run both models with the same setup, and would expect probabilities from \textsc{NorMistral} to still reflect correct answers.

In the 2-class setup, only responses that map to the binary classes were iterated through. We could then calculate the macro \fen score across the dataset for each likelihood, for each prompt. We use the likelihood method that gives the overall best \fen score for all prompts in the test set to evaluate which prompts we use for the few shot setup. %The output for a given answer looks like this: ["Det er positivt.",[0.001,0.526,0.836]], which would then be mapped to the \textsc{pos} class. 
We then select the best-performing prompts from each model.  %We select six sentences from each model: the two best 4-class and 2-binary prompts from the 4-class run, and the two best prompts from the 2-class. This is to make sure that there is at least two prompts that mention all classes.

\subsection{Zero-Shot Runs} 
In the zero-shot setup, we provide the model with a prompt, and calculate the likelihood for the 16 possible answers, using the likelihood estimates described above. Results for zero-shot runs with the two models, for both the 4-class and 2-class results, are reported in \cref{tab:probs}. We observe notable differences both between the two models, and between the binary and 4-classes. The \textsc{ChatNorT5} model performs much better in the 4-class setting, but this is mainly due to high scores for the negative and positive classes. We find that neutral and mixed are difficult for both models.

% \begin{table}[]
%     \centering
%     \begin{tabular}{@{}lrrrr@{}}
%     \toprule 
%     \textbf{Likelihood} & \textbf{CNT5 4} & \textbf{CNT5 2} & \textbf{MIS 4} & \textbf{MIS 2}\\
%     \midrule 
%          $\mathcal{L}_1$ & 0.399  & 0.887 & \textbf{0.094} & 0.848 \\
%          $\mathcal{L}_2$ & 0.406 & 0.892 & 0.028 & 0.890 \\
%          $\mathcal{L}_3$ & \textbf{0.424} & \textbf{0.893} & 0.027 & \textbf{0.891} \\
%     \bottomrule 
%     \end{tabular}
%     \caption{Highest values for the 4-class and 2-class results for \textsc{ChatNorT5} and \textsc{NorMistral} runs for the three likelihoods.}
%     \label{tab:probs}
% \end{table}

\begin{comment}

\begin{table}[]
    \centering
    \begin{tabular}{@{}lrr@{}}
    \toprule 
    \textbf{4-class scores} & \textbf{\textsc{ChatNorT5}} & \textbf{\textsc{NorMistral}} \\
    \midrule 
         $\mathcal{L}_1$ & 39.9  & \textbf{9.4} \\
         $\mathcal{L}_2$ & 40.6 & 2.8 \\
         $\mathcal{L}_3$ & \textbf{42.4} & 2.7 \\
    \bottomrule \\
    \\
    \toprule 
    \textbf{2-class scores} & \textbf{\textsc{ChatNorT5}} & \textbf{\textsc{NorMistral}}\\
    \midrule 
         $\mathcal{L}_1$ & 88.7 & 84.8 \\
         $\mathcal{L}_2$ & 89.2 & 89.0 \\
         $\mathcal{L}_3$ & \textbf{89.3} & \textbf{89.1} \\
    \bottomrule 
    \end{tabular}
    \caption{Zero-shot results. The highest F\textsubscript1 scores (among different prompts) for the 4-class and 2-class evaluation using the \textsc{ChatNorT5} and \textsc{NorMistral} models, and the three formulations of likelihood scores.}
    \label{tab:probs}
\end{table}

\end{comment}

\begin{table}[]
    \centering
    \begin{tabular}{@{}clrr@{}}
    \toprule 
    &  $\mathcal{L}$ & \textsc{ChatNorT5} & \textsc{NorMistral} \\
     \cmidrule(l){2-4} 
{\multirow{3}{*}{\rotatebox[origin=c]{90}{4-class}}}
   & $\mathcal{L}_1$ & 39.9          & \textbf{9.4} \\
   & $\mathcal{L}_2$ & 40.6          & 2.8 \\
   & $\mathcal{L}_3$ & \textbf{42.4} & 2.7 \\
    \midrule 
{\multirow{3}{*}{\rotatebox[origin=c]{90}{2-class}}}
   & $\mathcal{L}_1$ & 88.7          & 84.8 \\
   & $\mathcal{L}_2$ & 89.2          & 89.0 \\
   & $\mathcal{L}_3$ & \textbf{89.3} & \textbf{89.1} \\
    \bottomrule 
    \end{tabular}
    \caption{Zero-shot results. The highest F\textsubscript1 scores (among different prompts) for the 4-class and 2-class evaluation using the \textsc{ChatNorT5} and \textsc{NorMistral} models, and the three formulations of likelihood scores.}
    \label{tab:probs}
\end{table}

\paragraph{\textsc{ChatNorT5}}

We see that for \textsc{ChatNorT5}, it is the third likelihood that gives the best results, both in the 4-class and 2-class setup. The differences between the three values are not large, and we observe that the difference between the various prompts are far more marked. The two best overall prompts were \texttt{7-2-2} and \texttt{7-1-2}, and the best 4-class prompts were \texttt{7-2-4-2} and \texttt{8-2-4}. \texttt{7-2-2} was also the best binary prompt, along with \texttt{5-1-2}. We see that three of these are based on prompt 7, which is originally binary but expanded to 4-class in \texttt{7-2-4-2}. Regarding placement of the input sentence, in 19 of 24 pairs of sentences, having the sentence in front gives a higher macro \fen. In general, there is a problem that many prompts lead to few predictions of the neutral and mixed classes. %Two-class prompts invariably lead to no predictions in the neutral class. 
We also see that, in terms of difference between the likelihoods, invariably, if a prompt predicts overall more positive sentiment for one likelihood, it does so for all likelihoods. It seems that prompts that work well for the positive and negative classes outperform other prompts even if they predict more neutral and mixed classes. %Some of the high prediction of positive and negative, and less on neutral and mixed, could be due to the models earlier exposure to binary SA. 

\paragraph{\textsc{NorMistral}}
For \textsc{NorMistral} we see that all prompts almost exclusively predict the \textsc{mixed} class. Here, the difference is larger between the three likelihood scores. While the likelihood score $\mathcal{L}_1$ gives the best 4-class score, the best binary score is obtained again by $\mathcal{L}_3$. While $\mathcal{L}_3$ has no predictions outside the mixed class, $\mathcal{L}_2$ has a single prediction outside, but $\mathcal{L}_1$ has 201 positive, 416 negative, and 5612 neutral predictions. %A 7-based prompt, prompt \texttt{7-1-2-2}, is the prompt with the largest number of predictions in the \textsc{pos} and \textsc{neg} classes. 
The weak 4-class results are somewhat surprising, and can indicate either that the current prompts are largely inadequate for this model or that the model does not understand more nuanced sentiment. However, in the binary setup, the opposite is true. We approach very high \fen scores, suggesting that weak 4-class scores are a result of inadequate prompting. Interestingly, the best-performing sentences are four-class prompts: \texttt{9-1-4} and \texttt{6-1-4}.

Due to low and largely similar results for the 4-class setup, we select only four prompts for \textsc{NorMistral}, exclusively from the binary setup, and also include prompts \texttt{2-1-0} and \texttt{6-1-2-2}.

\subsection{Few-Shot Runs}

Having run zero-shot runs for both models, we use the best performing prompts in a four-shot setting. This setup consists of four pairs of query--response examples, one from each class, given to the model as context, before the test sentence we want to make a prediction for. Each example was randomly sampled from the training set, and selected separately for each sentence in the dataset. These examples are all taken from the training set. The best scores for the three likelihoods are reported in Table~\ref{tab:fewshot_probs}.

\begin{table}[]
    \centering
    \begin{tabular}{@{}clrr@{}}
    \toprule 
    &  $\mathcal{L}$ & \textsc{ChatNorT5} & \textsc{NorMistral} \\
     \cmidrule(l){2-4} 
{\multirow{3}{*}{\rotatebox[origin=c]{90}{4-class}}}
         & $\mathcal{L}_1$ & 28.6  & \textbf{12.2} \\
         & $\mathcal{L}_2$ & 28.6 & 2.7 \\
         & $\mathcal{L}_3$ & 28.6 & 2.7 \\
         \midrule 
{\multirow{3}{*}{\rotatebox[origin=c]{90}{2-class}}}         
         & $\mathcal{L}_1$ & 89.1 & \textbf{84.9} \\
         & $\mathcal{L}_2$ & \textbf{89.3} &83.9 \\
         & $\mathcal{L}_3$ & 83.6 & 83.9 \\
    \bottomrule 
    \end{tabular}
    \caption{Few-shot results. The highest \fen scores for the 4-class and 2-class, as for the zero-shot results.}
    \label{tab:fewshot_probs}
\end{table}

%Overall, we see that scores are low for the 4-class setting.
%The 2 x 2 best 2-class and 4-class predicting prompts from the zero-shot were selected, and run in a few-shot setup. For each N-shot run, we ran 4 x N sentences, where each sentence represents a separate class. In this way, the models is exposed to exactly one example of each class before we calculate the likelihoods. 

%For few-shot learning we select two main methods. The first method is to randomly sample a set of four examples for each sentence from the training set. The other method is to select a set of 4-example selections, that are run on the entire prompts. This more truly simulates a low resource setting, where we can measure the effect of only having access to a few examples. 

\paragraph{Few-Shot with \textsc{ChatNorT5}}
The results from the few-shot runs with \textsc{ChatNorT5} are reported in \cref{tab:few_shot_chatnort5}. We observe that the \fen score drops due to the 7-based prompts almost exclusively predicting the mixed class, while the 8-based prompt favors the neutral class, but is more balanced. In the binary setup, however, we see very high scores, almost beating the zero-shot results.%Should perhaps do a binary check here as well, but then the tables become really big.

\begin{table}[]
    \centering
    \begin{tabular}{@{}lrr@{}}
    \toprule
%     & \multicolumn{2}{c}{\textbf{Few-shot}}  \\ %& \multicolumn{2}{c}{2-shot}
%     \cmidrule(l){2-3}
     %\cmidrule(lr){4-5}
    \textbf{Prompt ID} & \textbf{4-class} & \textbf{2-class} \\ %& 4-class & 2-class \\
    \midrule
         \texttt{7-2-2}   & 3.8  (0.2)& 89.0 (0.3) \\%& 0.0  & 0.0 \\
         \texttt{7-1-2}   & 3.2 (0.1)& 85.2 (0.3)  \\%& 0.0  & 0.0\\
         \texttt{7-2-4-2} & 7.2 (0.2) & 87.8 (0.4) \\%& 0.0 & 0.0   \\
         \texttt{8-2-4}   & 26.1 (0.3) & 83.0 (0.3) \\%& 0.0 & 0.0  \\
        \texttt{5-1-2}   & 28.3 (0.2) & 62.0 (0.4) \\%& 0.0 & 0.0  \\
    \bottomrule
    \end{tabular}
    \caption{Mean \fen and standard deviation for the few-shot experiments using \textsc{ChatNorT5}. }
    \label{tab:few_shot_chatnort5}
\end{table}

\paragraph{Few-Shot with \textsc{NorMistral}}
The results of the four prompts from the \textsc{NorMistral} zero-shot run are reported in \cref{tab:few_shot_mistral}. We see that \textsc{NorMistral} also struggles with making reliable predictions in the 4-class setup, but performs well on \textsc{pos} and \textsc{neg}, albeit not as well as \textsc{ChatNorT5}.

\begin{table}[]
    \centering
    \begin{tabular}{@{}lrr@{}}
    \toprule
%     & \multicolumn{2}{c}{\textbf{Few-shot}}  \\ %& \multicolumn{2}{c}{2-shot}
%     \cmidrule(l){2-3} 
     %\cmidrule(lr){4-5}
    \textbf{Prompt ID} & \textbf{4-class} & \textbf{2-class} \\ %& 4-class & 2-class \\
    \midrule
         \texttt{2-1-0}   & 6.5 (0.4)& 84.4 (0.1) \\%& 0.0  & 0.0 \\
         \texttt{6-1-4}   & 11.8 (0.2)& 76.4 (0.5)  \\%& 0.0  & 0.0\\
         \texttt{6-1-2-2} & 11.8 (0.2) & 72.3 (0.3) \\%& 0.0 & 0.0   \\
         \texttt{9-1-4}   & 11.0 (0.1) & 69.5 (0.5) \\%& 0.0 & 0.0  \\
    \bottomrule
    \end{tabular}
    \caption{Mean \fen and standard deviations for the few-shot runs using \textsc{NorMistral}. }
    \label{tab:few_shot_mistral}
\end{table}

\paragraph{Comparison with Baseline}
We observe that for the 4-class setup, \textsc{ChatNorT5} achieves similar scores to the baseline model in the zero-shot runs, while \textsc{NorMistral} achieves notably lower scores for all runs using the 4-class setup. However, for binary sentiment, both generative models achieve higher scores than the baseline in most cases, both for zero-shot and few-shot. The low scores in the 4-class setup are surprising for both models, and we hope to investigate this in later experiments.

\subsection{Comparison with Human Annotators}

%\subsection{Best Model and Prompt combinations}
Due to the inability of any prompt to reliably predict the \textsc{neut} and \textsc{mix} classes, we selected prompts based on the best 2-class results, along with the best four-class and binary for zero shot \textsc{ChatNorT5}. This gives us three prompts: \texttt{7-2-2}, \texttt{2-1-0} and \texttt{9-1-4}.

\paragraph{Model-annotator IAA}
We compared these three combinations of prompts with models, comparing them individually with each annotator, which have the results presented in \cref{tab:annotator_iaa}. We treat all documents annotated by an annotator as representing that annotator, and calculate kappa scores between the model and the human annotators, like we did with the human annotators. For the full 4-class problem, we see that agreement is low, but still not as low as some of the project-initial disagreements.

\begin{table}[]
    \centering
    \begin{tabular}{@{}lrrrrr@{}}
    \toprule
            \textbf{Prompt ID} & \textbf{A1} & \textbf{A2} & \textbf{A4} & \textbf{A6} & \textbf{A7}  \\
               \midrule
    \texttt{2-1-0} (2)  &  0.69 & 0.00 & 0.70 & 0.62 & 0.70\\
    \texttt{7-2-2} (4)  &  0.46 & 0.32 & 0.52 & 0.49 & 0.55\\
    \texttt{7-2-2} (2) &  0.79 & 0.00 & 0.81 & 0.76 & 0.83\\
    \texttt{9-1-4} (2) &  0.81 & 0.00 & 0.77 & 0.75 & 0.83\\
    \bottomrule
    \end{tabular}
    \caption{Annotation agreement between the best predictions and the human annotators. Annotators A3 and A5 are not represented due to lack of applicable data in the test set. Prompt 7-2-2 is tested both in the 2-class and the 4-class setup. Annotator A7 did not partake in the previous IAA rounds.}
    \label{tab:annotator_iaa}
\end{table}

\paragraph{Human versus Model}

While we found the \fen scores to much higher for the 2-class setup, treating the model outputs as annotations still gives us lower scores than that expected from humans familiar with the task. The obvious issue is that we struggle to get our models to reliably distinguish between neutral and polar sentences. Inspection shows that a prevalent error for both prompt \texttt{2-1-0} and \texttt{7-2-2} is to mistake positive sentences for negative, while prompt \texttt{9-1-4} has more cases where the model treats positive as negative. The most common mistake in the 4-class setup for \texttt{7-2-2} is that non-mixed sentences are classified as mixed. 

\section{Conclusion}

This paper has described how free-text comments in patient surveys collected by the Norwegian Institute of Public Health have been annotated with information about sentiment. Specifically, our data comprise patient comments in Norwegian on experiences with General Practitioners and Special Mental Healthcare, which we have annotated with positive/negative polarity (including intensity) on both the comment- and sentence-level. In addition to describing the annotation guidelines and presenting an analysis of the resulting dataset, we also include experimental results on augmenting the human annotations with predictions by two different open-source pretrained large language models (LLMs); \textsc{ChatNorT5} and \textsc{NorMistral}. We report results for both zero- and few-shot settings for several different prompting configurations. We find that the predictions of the LLMs are sensitive to the particular prompt used, and that the best configuration depends on the specific model. Moreover, while we find that both models perform well for the simple binary cases where sentences are either positive or negative, they both struggle with neutral and mixed-polarity examples. Our error analysis shows that the predictions of the LMMs used in this study are still inferior in quality to the human annotations for our dataset.

\section{Limitations}% and ethical considerations}

\paragraph{Annotator representations}
Due to work load limitations, we were not able to provide an even distribution of data across human annotators. This makes the claims on some of the annotators hold less than for others. 

\paragraph{Intensity}
While we would have liked to include intensity scores, this will have to be the subject for later research. While interesting due to being a source of disagreement in humans, and we believe that differences in the treatment of intensity might reveal further differences between humans and models, it requires more space than what we could dedicate in this paper. 

\paragraph{Variation}
We note that there is linguistic variation in the dataset, but addressing this is outside the scope of our paper. We hope to be able to return to this to be able to better assess how user language might affect how patients' voices are analyzed using systems often trained and evaluated on normative and edited language.

\section{Ethical Considerations}
%\paragraph{Commercial models}
While it might be possible to get similar or even better results with certain commercial models, there are several reasons why we opt for open-source Norwegian models. First of all, these models can be run locally, also without using APIs that would require sending data to servers not cleared for storage of our data, and do not pose any conflict in terms of privacy or potential data leakage. Secondly, models trained on data from the same area as the patients might lead to less likelihood of cultural bias affecting judgements. Finally, the models' training data are open, and it is therefore possible to investigate biases and potential problems should they arise.

%\paragraph{Training data}
%ChatNorT5 was trained on translated material. As we do not have comparable material in manually translated Norwegian, it is difficult to assess the effect this has on the model, and how it influences this task. We observe that there are clear traces of translationese, but that the model still produces grammatical Norwegian in most cases.

\section{Acknowledgements}
We would like to thank Karoline Aasgaard, Kirsten Danielsen, Lina Harvold Ellingsen-Dalskau, Hilde Hestad Iversen and Inger Opedal Paulsrud for their annotation work. This work was supported by the Norwegian Research Council, through the project \textit{Styrking av pasientstemmen i helsetjenesteevaluering: maskinlæring på fritekstkommentarer fra spørreundersøkelser og nettkilder.} `Strengthening the patient voice in health service evaluation: machine learning on free-text comments from surveys and online sources', project number 103707. The computations were performed on resources provided through Sigma2 -- the national research infrastructure provider for High-Performance Computing and large-scale data storage in Norway.

%\subsection{Training and Hardware}
%training time: The project utilized the educloud 
%This involves implicitly concluding that humans actually have the gold data

%Is it possible to simulate mixed sentiment by setting a threshold for certain polarities, and then predict positive and negative separately? ANd then maybe say that whatever has a certain low threshold for pos and neg must be neut? Check the likelihoods against the classes!

\nocite{*}
\section{Bibliographical References}\label{sec:reference}

\bibliographystyle{lrec-coling2024-natbib}
\bibliography{lrec-coling2024-example}

\begin{thebibliography}{42}
\expandafter\ifx\csname natexlab\endcsname\relax\def\natexlab#1{#1}\fi

\bibitem[{Alemi et~al.(2012)Alemi, Torii, Clementz, and Aron}]{alemi2012}
Farrokh Alemi, Manabu Torii, Laura Clementz, and David~C. Aron. 2012.
\newblock \href {https://journals.lww.com/qmhcjournal/fulltext/2012/01000/feasibility_of_real_time_satisfaction_surveys.4.aspx} {Feasibility of real-time satisfaction surveys through automated analysis of patients' unstructured comments and sentiments}.
\newblock \emph{Quality Management in Healthcare}, 21(1).

\bibitem[{Alexander et~al.(2022)Alexander, Bahja, and Butt}]{alexander2022}
George Alexander, Mohammed Bahja, and Gibran~Farook Butt. 2022.
\newblock \href {https://doi.org/10.2196/29385} {Automating large-scale health care service feedback analysis: Sentiment analysis and topic modeling study}.
\newblock \emph{JMIR Med Inform}, 10(4):e29385.

\bibitem[{Anhang~Price et~al.(2014)Anhang~Price, Elliott, Zaslavsky, Hays, Lehrman, Rybowski, Edgman-Levitan, and Cleary}]{Anhang_Price2014-ln}
Rebecca Anhang~Price, Marc~N Elliott, Alan~M Zaslavsky, Ron~D Hays, William~G Lehrman, Lise Rybowski, Susan Edgman-Levitan, and Paul~D Cleary. 2014.
\newblock Examining the role of patient experience surveys in measuring health care quality.
\newblock \emph{Med Care Res Rev}, 71(5):522--554.

\bibitem[{Bahja and Lycett(2016)}]{bahja2016}
Mohammed Bahja and Mark Lycett. 2016.
\newblock Identifying patient experience from online resources via sentiment analysis and topic modelling.
\newblock In \emph{Proceedings of the 3rd {IEEE/ACM} International Conference on Big Data Computing, Applications and Technologies}, Shanghai, China.

\bibitem[{Bird et~al.(2009)Bird, Klein, and Loper}]{bird2009natural}
Steven Bird, Ewan Klein, and Edward Loper. 2009.
\newblock \emph{Natural language processing with Python: analyzing text with the natural language toolkit}.
\newblock O'Reilly Media, Inc.

\bibitem[{Brown et~al.(2020)Brown, Mann, Ryder, Subbiah, Kaplan, Dhariwal, Neelakantan, Shyam, Sastry, Askell et~al.}]{brown2020language}
Tom Brown, Benjamin Mann, Nick Ryder, Melanie Subbiah, Jared~D Kaplan, Prafulla Dhariwal, Arvind Neelakantan, Pranav Shyam, Girish Sastry, Amanda Askell, et~al. 2020.
\newblock Language models are few-shot learners.
\newblock \emph{Advances in neural information processing systems}, 33:1877--1901.

\bibitem[{Ding et~al.(2023)Ding, Chen, Xu, Qin, Zheng, Hu, Liu, Sun, and Zhou}]{ding2023enhancing}
Ning Ding, Yulin Chen, Bokai Xu, Yujia Qin, Zhi Zheng, Shengding Hu, Zhiyuan Liu, Maosong Sun, and Bowen Zhou. 2023.
\newblock Enhancing chat language models by scaling high-quality instructional conversations.
\newblock \emph{arXiv preprint arXiv:2305.14233}.

\bibitem[{Doing-Harris et~al.(2017)Doing-Harris, Mowery, Daniels, Chapman, and Conway}]{Doing-Harris2017-hq}
Kristina Doing-Harris, Danielle~L Mowery, Chrissy Daniels, Wendy~W Chapman, and Mike Conway. 2017.
\newblock Understanding patient satisfaction with received healthcare services: A natural language processing approach.
\newblock \emph{AMIA Annu Symp Proc}, 2016:524--533.

\bibitem[{Doyle et~al.(2013)Doyle, Lennox, and Bell}]{Doylee001570}
Cathal Doyle, Laura Lennox, and Derek Bell. 2013.
\newblock \href {https://doi.org/10.1136/bmjopen-2012-001570} {A systematic review of evidence on the links between patient experience and clinical safety and effectiveness}.
\newblock \emph{BMJ Open}, 3(1).

\bibitem[{Go et~al.(2009)Go, Bhayani, and Huang}]{sentiment140}
Alec Go, Richa Bhayani, and Lei Huang. 2009.
\newblock Twitter sentiment classification using distant supervision.
\newblock Technical report, CS224N project report, Stanford.

\bibitem[{Greaves et~al.(2013)Greaves, Ramirez-Cano, Millett, Darzi, and Donaldson}]{greaves2013}
Felix Greaves, Daniel Ramirez-Cano, Christopher Millett, Ara Darzi, and Liam Donaldson. 2013.
\newblock \href {https://doi.org/10.2196/jmir.2721} {Use of sentiment analysis for capturing patient experience from free-text comments posted online}.
\newblock \emph{J Med Internet Res}, 15(11):e239.

\bibitem[{Grob et~al.(2019{\natexlab{a}})Grob, Schlesinger, Barre, Bardach, Lagu, Shaller, Parker, Martino, Finucane, Cerully, and Palimaru}]{article}
Rachel Grob, Mark Schlesinger, Lacey Barre, Naomi Bardach, Tara Lagu, Dale Shaller, Andrew Parker, Steven Martino, Melissa Finucane, Jennifer Cerully, and Alina Palimaru. 2019{\natexlab{a}}.
\newblock \href {https://doi.org/10.1111/1468-0009.12374} {What words convey: The potential for patient narratives to inform quality improvement}.
\newblock \emph{The Milbank Quarterly}, 97:176--227.

\bibitem[{Grob et~al.(2019{\natexlab{b}})Grob, Schlesinger, Barre, Bardach, Lagu, Shaller, Parker, Martino, Finucane, Cerully, and Palimaru}]{Grob2019-do}
Rachel Grob, Mark Schlesinger, Lacey~Rose Barre, Naomi Bardach, Tara Lagu, Dale Shaller, Andrew~M Parker, Steven~C Martino, Melissa~L Finucane, Jennifer~L Cerully, and Alina Palimaru. 2019{\natexlab{b}}.
\newblock What words convey: The potential for patient narratives to inform quality improvement.
\newblock \emph{The Milbank Quarterly}, 97(1):176--227.

\bibitem[{Hawkins et~al.(2016)Hawkins, Brownstein, Tuli, Runels, Broecker, Nsoesie, McIver, Rozenblum, Wright, Bourgeois, and Greaves}]{Hawkins404}
Jared~B Hawkins, John~S Brownstein, Gaurav Tuli, Tessa Runels, Katherine Broecker, Elaine~O Nsoesie, David~J McIver, Ronen Rozenblum, Adam Wright, Florence~T Bourgeois, and Felix Greaves. 2016.
\newblock \href {https://doi.org/10.1136/bmjqs-2015-004309} {Measuring patient-perceived quality of care in us hospitals using twitter}.
\newblock \emph{BMJ Quality \& Safety}, 25(6):404--413.

\bibitem[{Huppertz and Otto(2018)}]{huppertz}
John~W. Huppertz and Peter Otto. 2018.
\newblock \href {https://journals.lww.com/hcmrjournal/fulltext/2018/10000/predicting_hcahps_scores_from_hospitals__social.10.aspx} {Predicting hcahps scores from hospitals' social media pages: A sentiment analysis}.
\newblock \emph{Health Care Management Review}, 43(4).

\bibitem[{Iversen et~al.(2014)Iversen, Bjertn{\ae}s, and Skudal}]{iversen}
Hilde Iversen, {\O}yvind Bjertn{\ae}s, and Kjersti Skudal. 2014.
\newblock \href {https://doi.org/10.1136/bmjopen-2014-004848} {Patient evaluation of hospital outcomes: An analysis of open-ended comments from extreme clusters in a national survey}.
\newblock \emph{BMJ open}, 4:e004848.

\bibitem[{Iversen et~al.(2022)Iversen, Haugum, and Bjertn{\ae}s}]{Iversen2022-ju}
Hilde~Hestad Iversen, Mona Haugum, and {\O}yvind Bjertn{\ae}s. 2022.
\newblock Reliability and validity of the psychiatric inpatient patient experience questionnaire -- continuous electronic measurement ({PIPEQ-CEM}).
\newblock \emph{BMC Health Services Research}, 22(1):897.

\bibitem[{Jiang et~al.(2023)Jiang, Sablayrolles, Mensch, Bamford, Chaplot, de~las Casas, Bressand, Lengyel, Lample, Saulnier, Lavaud, Lachaux, Stock, Scao, Lavril, Wang, Lacroix, and Sayed}]{jiang2023mistral}
Albert~Q. Jiang, Alexandre Sablayrolles, Arthur Mensch, Chris Bamford, Devendra~Singh Chaplot, Diego de~las Casas, Florian Bressand, Gianna Lengyel, Guillaume Lample, Lucile Saulnier, Lélio~Renard Lavaud, Marie-Anne Lachaux, Pierre Stock, Teven~Le Scao, Thibaut Lavril, Thomas Wang, Timothée Lacroix, and William~El Sayed. 2023.
\newblock \href {http://arxiv.org/abs/2310.06825} {Mistral 7b}.

\bibitem[{Jiménez~Zafra et~al.(2017)Jiménez~Zafra, Martín~Valdivia, Maks, and Izquierdo~Beviá}]{zafra}
Salud~M. Jiménez~Zafra, María~Teresa Martín~Valdivia, Isa Maks, and Rubén Izquierdo~Beviá. 2017.
\newblock {Analysis of patient satisfaction in Dutch and Spanish online reviews}.

\bibitem[{Khanbhai et~al.(2021)Khanbhai, Anyadi, Symons, Flott, Darzi, and Mayer}]{Khanbhai2021}
Mustafa Khanbhai, Patrick Anyadi, Joshua Symons, Kelsey Flott, Ara Darzi, and Erik Mayer. 2021.
\newblock \href {https://doi.org/10.1136/bmjhci-2020-100262} {Applying natural language processing and machine learning techniques to patient experience feedback: a systematic review}.
\newblock \emph{BMJ Health \& Care Informatics}, 28(1).

\bibitem[{Khanbhai et~al.(2022)Khanbhai, Warren, Symons, Flott, Harrison-White, Manton, Darzi, and Mayer}]{khanbai2022}
Mustafa Khanbhai, Leigh Warren, Joshua Symons, Kelsey Flott, Stephanie Harrison-White, Dave Manton, Ara Darzi, and Erik Mayer. 2022.
\newblock \href {https://doi.org/https://doi.org/10.1016/j.ijmedinf.2021.104642} {Using natural language processing to understand, facilitate and maintain continuity in patient experience across transitions of care}.
\newblock \emph{International Journal of Medical Informatics}, 157:104642.

\bibitem[{Kim et~al.(2022)Kim, Hessel, Jiang, West, Lu, Yu, Zhou, Bras, Alikhani, Kim, Sap, and Choi}]{kim2022soda}
Hyunwoo Kim, Jack Hessel, Liwei Jiang, Peter West, Ximing Lu, Youngjae Yu, Pei Zhou, Ronan~Le Bras, Malihe Alikhani, Gunhee Kim, Maarten Sap, and Yejin Choi. 2022.
\newblock Soda: Million-scale dialogue distillation with social commonsense contextualization.
\newblock \emph{ArXiv}, abs/2212.10465.

\bibitem[{Kj{\o}llesdal et~al.(2020)Kj{\o}llesdal, Indseth, Iversen, and Bjertn{\ae}s}]{Kjollesdal2020-wz}
Marte Kj{\o}llesdal, Thor Indseth, Hilde~Hestad Iversen, and {\O}yvind Bjertn{\ae}s. 2020.
\newblock Patient experiences with general practice in norway: a comparison of immigrant groups and the majority population following a national survey.
\newblock \emph{BMC Health Serv Res}, 20(1):1106.

\bibitem[{Kolesnichenko et~al.(2023)Kolesnichenko, Velldal, and {\O}vrelid}]{kolesnichenko-etal-2023-word}
Larisa Kolesnichenko, Erik Velldal, and Lilja {\O}vrelid. 2023.
\newblock \href {https://aclanthology.org/2023.resourceful-1.6} {Word substitution with masked language models as data augmentation for sentiment analysis}.
\newblock In \emph{Proceedings of the Second Workshop on Resources and Representations for Under-Resourced Languages and Domains (RESOURCEFUL-2023)}, pages 42--47, T{\'o}rshavn, the Faroe Islands. Association for Computational Linguistics.

\bibitem[{Maas et~al.(2011)Maas, Daly, Pham, Huang, Ng, and Potts}]{maas-etal-2011-learning}
Andrew~L. Maas, Raymond~E. Daly, Peter~T. Pham, Dan Huang, Andrew~Y. Ng, and Christopher Potts. 2011.
\newblock \href {https://aclanthology.org/P11-1015} {Learning word vectors for sentiment analysis}.
\newblock In \emph{Proceedings of the 49th Annual Meeting of the Association for Computational Linguistics: Human Language Technologies}, pages 142--150, Portland, Oregon, USA. Association for Computational Linguistics.

\bibitem[{M{\ae}hlum et~al.(2019)M{\ae}hlum, Barnes, {\O}vrelid, and Velldal}]{maehlum-etal-2019-annotating}
Petter M{\ae}hlum, Jeremy Barnes, Lilja {\O}vrelid, and Erik Velldal. 2019.
\newblock \href {https://www.aclweb.org/anthology/W19-6113} {Annotating evaluative sentences for sentiment analysis: a dataset for {N}orwegian}.
\newblock In \emph{Proceedings of the 22nd Nordic Conference on Computational Linguistics}, pages 121--130, Turku, Finland. Link{\"o}ping University Electronic Press.

\bibitem[{{\O}vrelid et~al.(2020){\O}vrelid, M{\ae}hlum, Barnes, and Velldal}]{ovrelid-etal-2020-fine}
Lilja {\O}vrelid, Petter M{\ae}hlum, Jeremy Barnes, and Erik Velldal. 2020.
\newblock \href {https://aclanthology.org/2020.lrec-1.618} {A fine-grained sentiment dataset for {N}orwegian}.
\newblock In \emph{Proceedings of the Twelfth Language Resources and Evaluation Conference}, pages 5025--5033, Marseille, France. European Language Resources Association.

\bibitem[{Qi et~al.(2020)Qi, Zhang, Zhang, Bolton, and Manning}]{qi2020stanza}
Peng Qi, Yuhao Zhang, Yuhui Zhang, Jason Bolton, and Christopher~D. Manning. 2020.
\newblock Stanza: A python natural language processing toolkit for many human languages.

\bibitem[{Raffel et~al.(2020)Raffel, Shazeer, Roberts, Lee, Narang, Matena, Zhou, Li, and Liu}]{10.5555/3455716.3455856}
Colin Raffel, Noam Shazeer, Adam Roberts, Katherine Lee, Sharan Narang, Michael Matena, Yanqi Zhou, Wei Li, and Peter~J. Liu. 2020.
\newblock Exploring the limits of transfer learning with a unified text-to-text transformer.
\newblock \emph{J. Mach. Learn. Res.}, 21(1).

\bibitem[{Riiskj{\ae}r et~al.(2012)Riiskj{\ae}r, Ammentorp, and Kofoed}]{Riiskjaer2012-uh}
Erik Riiskj{\ae}r, Jette Ammentorp, and Poul-Erik Kofoed. 2012.
\newblock The value of open-ended questions in surveys on patient experience: number of comments and perceived usefulness from hospital perspective.
\newblock \emph{International Journal for Quality in Health Care}, 24(5):509--516.

\bibitem[{Rivas et~al.(2019)Rivas, Tkacz, Antao, Mentzakis, Gordon, Anstee, and Giordano}]{rivas2019automated}
Carol Rivas, Daria Tkacz, Laurence Antao, Emmanouil Mentzakis, Margaret Gordon, Sydney Anstee, and Richard Giordano. 2019.
\newblock Automated analysis of free-text comments and dashboard representations in patient experience surveys: a multimethod co-design study.
\newblock \emph{Health Services and Delivery Research}, (7.23).

\bibitem[{Rohde et~al.(2022)Rohde, Mashinchi, Rizun, Gruda, Foley, Flynn, and Ojo}]{Rohde2022}
Daniela Rohde, Mona~Isazad Mashinchi, Nina Rizun, Dritjon Gruda, Conor Foley, Rachel Flynn, and Adegboyega Ojo. 2022.
\newblock \href {http://arxiv.org/abs/https://qualitysafety.bmj.com/content/25/8/604.full.pdf} {Generating actionable insights from free-text care experience survey data using qualitative and computational text analysis: A study protocol}.
\newblock \emph{HRB Open Research}.

\bibitem[{R{\o}nningstad et~al.(2022)R{\o}nningstad, Velldal, and {\O}vrelid}]{ronningstad-etal-2022-entity}
Egil R{\o}nningstad, Erik Velldal, and Lilja {\O}vrelid. 2022.
\newblock \href {https://aclanthology.org/2022.coling-1.589} {Entity-level sentiment analysis ({ELSA}): An exploratory task survey}.
\newblock In \emph{Proceedings of the 29th International Conference on Computational Linguistics}, pages 6773--6783, Gyeongju, Republic of Korea. International Committee on Computational Linguistics.

\bibitem[{Samuel et~al.(2023)Samuel, Kutuzov, Touileb, Velldal, {\O}vrelid, R{\o}nningstad, Sigdel, and Palatkina}]{samuel-etal-2023-norbench}
David Samuel, Andrey Kutuzov, Samia Touileb, Erik Velldal, Lilja {\O}vrelid, Egil R{\o}nningstad, Elina Sigdel, and Anna Palatkina. 2023.
\newblock \href {https://aclanthology.org/2023.nodalida-1.61} {{N}or{B}ench {--} a benchmark for {N}orwegian language models}.
\newblock In \emph{Proceedings of the 24th Nordic Conference on Computational Linguistics (NoDaLiDa)}, pages 618--633, T{\'o}rshavn, Faroe Islands. University of Tartu Library.

\bibitem[{Socher et~al.(2013)Socher, Perelygin, Wu, Chuang, Manning, Ng, and Potts}]{socher-etal-2013-recursive}
Richard Socher, Alex Perelygin, Jean Wu, Jason Chuang, Christopher~D. Manning, Andrew Ng, and Christopher Potts. 2013.
\newblock \href {https://aclanthology.org/D13-1170} {Recursive deep models for semantic compositionality over a sentiment treebank}.
\newblock In \emph{Proceedings of the 2013 Conference on Empirical Methods in Natural Language Processing}, pages 1631--1642, Seattle, Washington, USA. Association for Computational Linguistics.

\bibitem[{Taori et~al.(2023)Taori, Gulrajani, Zhang, Dubois, Li, Guestrin, Liang, and Hashimoto}]{taori2023alpaca}
Rohan Taori, Ishaan Gulrajani, Tianyi Zhang, Yann Dubois, Xuechen Li, Carlos Guestrin, Percy Liang, and Tatsunori~B Hashimoto. 2023.
\newblock Alpaca: A strong, replicable instruction-following model.
\newblock \emph{Stanford Center for Research on Foundation Models. https://crfm. stanford. edu/2023/03/13/alpaca. html}, 3(6):7.

\bibitem[{Tiedemann and Thottingal(2020)}]{TiedemannThottingal:EAMT2020}
J{\"o}rg Tiedemann and Santhosh Thottingal. 2020.
\newblock {OPUS-MT} — {B}uilding open translation services for the {W}orld.
\newblock In \emph{Proceedings of the 22nd Annual Conferenec of the European Association for Machine Translation (EAMT)}, Lisbon, Portugal.

\bibitem[{van Buchem et~al.(2022)van Buchem, Neve, Kant, Steyerberg, Boosman, and Hensen}]{vanBuchem2022}
Marieke~M. van Buchem, Olaf~M. Neve, Ilse M.~J. Kant, Ewout~W. Steyerberg, Hileen Boosman, and Erik~F. Hensen. 2022.
\newblock \href {https://doi.org/10.1186/s12911-022-01923-5} {Analyzing patient experiences using natural language processing: development and validation of the artificial intelligence patient reported experience measure (ai-prem)}.
\newblock \emph{BMC Medical Informatics and Decision Making}, 22(1):183.

\bibitem[{Velldal et~al.(2018)Velldal, {\O}vrelid, Bergem, Stadsnes, Touileb, and J{\o}rgensen}]{velldal-etal-2018-norec}
Erik Velldal, Lilja {\O}vrelid, Eivind~Alexander Bergem, Cathrine Stadsnes, Samia Touileb, and Fredrik J{\o}rgensen. 2018.
\newblock \href {https://aclanthology.org/L18-1661} {{N}o{R}e{C}: The {N}orwegian review corpus}.
\newblock In \emph{Proceedings of the Eleventh International Conference on Language Resources and Evaluation ({LREC} 2018)}, Miyazaki, Japan. European Language Resources Association (ELRA).

\bibitem[{Wagland et~al.(2016)Wagland, Recio-Saucedo, Simon, Bracher, Hunt, Foster, Downing, Glaser, and Corner}]{Wagland604}
Richard Wagland, Alejandra Recio-Saucedo, Michael Simon, Michael Bracher, Katherine Hunt, Claire Foster, Amy Downing, Adam Glaser, and Jessica Corner. 2016.
\newblock \href {https://doi.org/10.1136/bmjqs-2015-004063} {Development and testing of a text-mining approach to analyse patients{\textquoteright} comments on their experiences of colorectal cancer care}.
\newblock \emph{BMJ Quality \& Safety}, 25(8):604--614.

\bibitem[{Wei et~al.(2021)Wei, Bosma, Zhao, Guu, Yu, Lester, Du, Dai, and Le}]{weifinetuned}
Jason Wei, Maarten Bosma, Vincent Zhao, Kelvin Guu, Adams~Wei Yu, Brian Lester, Nan Du, Andrew~M Dai, and Quoc~V Le. 2021.
\newblock Finetuned language models are zero-shot learners.
\newblock In \emph{International Conference on Learning Representations}.

\bibitem[{Wei et~al.(2022)Wei, Bosma, Zhao, Guu, Yu, Lester, Du, Dai, and Le}]{wei2022finetuned}
Jason Wei, Maarten Bosma, Vincent Zhao, Kelvin Guu, Adams~Wei Yu, Brian Lester, Nan Du, Andrew~M. Dai, and Quoc~V Le. 2022.
\newblock \href {https://openreview.net/forum?id=gEZrGCozdqR} {Finetuned language models are zero-shot learners}.
\newblock In \emph{Proceedings of the 10th International Conference on Learning Representations}, online.

\end{thebibliography}

%\section{Language Resource References}
%\label{lr:ref}
%\bibliographystylelanguageresource{lrec-coling2024-natbib}
%\bibliographylanguageresource{languageresource}

\end{document}